\documentclass{article}
\usepackage{PRIMEarxiv}
\usepackage{cite}
\usepackage{amsmath,amssymb,amsfonts}
\usepackage{algorithmic}
\usepackage{graphicx}
\usepackage{textcomp}
\usepackage{xcolor}
\usepackage{subfig}
\title{Multipod Convolutional Network\\
{\footnotesize}
}

\author{
  Hongyi Pan, Salih Atici, Ahmet Enis Cetin \\
  Department of Electrical and Computer Engineering\\
  University of Illinois Chicago\\
  Chicago, Illinois, USA\\
  \texttt{\{hpan21, satici2, aecyy\}@uic.edu} \\
}

\begin{document}

\maketitle

\begin{abstract}
In this paper, we introduce a convolutional network which we call MultiPodNet consisting of a combination of two or more convolutional networks which process the input image in parallel to achieve the same goal. Output feature maps of parallel convolutional networks are fused at the fully connected layer of the network. We experimentally observed that three parallel pod networks (TripodNet) produce the best results in commonly used object recognition datasets. Baseline pod networks can be of any type. In this paper, we use ResNets as baseline networks and their inputs are augmented image patches. The number of parameters of the TripodNet is about three times that of a single ResNet. We train the TripodNet using the standard backpropagation type algorithms. In each individual ResNet, parameters are initialized with different random numbers during training.
The TripodNet achieved state-of-the-art performance on CIFAR-10 and ImageNet datasets. For example, it improved the accuracy of a single ResNet from 91.66\% to 92.47\% under the same training process on the CIFAR-10 dataset.
\end{abstract}

\section{Introduction}

In this article, we describe a novel convolutional network which we call MultiPodNet consisting of a combination of two or more convolutional networks which process the input image or data in parallel.  Output feature maps of parallel convolutional networks are fused at the fully connected layer of the network.  In other words, we have a bank of networks consisting of two more networks forming the main body of the MultiPodNet. 

Recent studies involving convolutional neural networks and vision transformers have shown great success in image classification tasks and they have created different perspectives. The machine learning revolution that started with a deep convolutional network continues with hierarchical vision Transformer using shifted windows and next-generation convolutional neural networks  \cite{bengio1993globally}, \cite{lecun1995convolutional}, \cite{lecun2015deep}. The great success of transformers, Swin Transformers on image classification tasks showed that increasing the number of network parameters and using images as a series of patches help to increase the performance of the model \cite{vaswani2017attention}. Recently introduced ConvNeXt network \cite{liu2022convnet} which is also a huge convolutional network is as successful as transformer-type networks. It also has a huge number of parameters and it is trained using novel and different methods to improve the classical convolutional models. 

In this paper, we increase the number of parameters of the deep neural network using a bank of parallel networks working towards the same goal such as object recognition. The MultiPod network 
can also process the input image as image patches in parallel as in transformer networks. The original input image and/or its augmented versions are fed into convolutional networks forming the multipod network and the output feature maps of parallel convolutional networks are combined before the fully connected dense layer. Since the convolutional neural networks are used in parallel, the network is called MultiPodNet.
The difference between MultiPodNet and other concatenated networks is that the same input image is used as input to each of the networks forming the MultipodNet. We also use image patches and different augmentation techniques to create several instances of the original input. Depending on the image database  and the object recognition problem different augmentation methods and image patches can be used as input to the MultiPod Network. 

We initialize the parallel networks with different random numbers during the training process. This approach leads to different parameters in parallel networks and that is how we improve the recognition capability of the single network. Pod networks  capture the fundamental and structural information in the image more than a single network
. 
In this paper, we present experimental results and observed that the Tripod network with three parallel networks achieve the best results in CIFAR-10 dataset. The Tripod network consisting of three ResNet-20 performs better than a single ResNet-20. The Tripod network achieves a state-of-the-art performance on CIFAR-10 dataset, obtaining 92.47\%. This result is also better than the accuracy results of ViT and Swin Transformer networks. 



We first introduced the concept of two parallel networks working towards the same goal in \cite{deveci2018energy}. There were only two networks in \cite{deveci2018energy}. One of the networks has binary weights and convolutions in one of the networks was in the image domain and the second network was a neural network transforming the data to the Hadamard transform domain and processing it in the transform domain \cite{deveci2018energy,pan2021fast, pan2022deep, pan2022block}.
Other parallel networks include Siamese type networks and non-contrastive learning methods \cite{chen2021exploring,grill2020bootstrap}.
Xun Huang \textit{et al.} proposed a DNN with two parallel branches but one of the networks is trained with low-resolution images and the goal of the structure was salience map prediction~\cite{huang2015salicon}.

The organization of the paper is as follows. We describe the architecture of Multipod networks in Section~\ref{sec: Methodology} and present experimental results on the CIFAR-10 \cite{krizhevsky2010convolutional} and ImageNet-1K~\cite{deng2009imagenet} databases in Section~\ref{sec: Experimental}. Section~\ref{sec: Concolusion} describes the conclusions and future work.

\section{Methodology}\label{sec: Methodology}
In this section, we describe the structure of multipod networks along with the different training techniques we used in this study. 
Each convolutional network is trained separately and as a result, they have different convolutional weights. They produce different outputs for a given input image. The output feature maps of every network are combined and connected to fully connected dense layers in two different ways as shown in Fig. \ref{fig: resnet}. The network is used in image classification tasks and aims to achieve state-of-the-art performance. 

It is also possible to input the original image in the form of overlapping or non-overlapping patches and in augmented forms as in Swin transformers, BYOL, SimSiam, and ConvNext models. 

\subsection{Input to the Multipod Network}
It is stated that data augmentation boosts the performance of convolutional networks, and it helps to avoid overfitting. Although many pre-trained networks are trained on huge datasets such as Coco or ImageNet, recent studies make use of advanced data augmentation techniques such as Mixup~\cite{zhang2017mixup}, Cutmix~\cite{yun2019cutmix}, RandAugment~\cite{cubuk2020randaugment} to improve the performance. 

\begin{figure}[htbp]
		\begin{center}
			\subfloat[]{\includegraphics[width=0.1\linewidth]{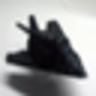}}
			\subfloat[]{\includegraphics[width=0.1\linewidth]{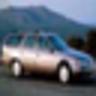}}
			\subfloat[]{\includegraphics[width=0.1\linewidth]{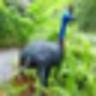}}
			\subfloat[]{\includegraphics[width=0.1\linewidth]{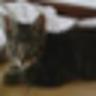}}
			\subfloat[]{\includegraphics[width=0.1\linewidth]{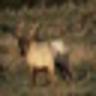}}
			\subfloat[]{\includegraphics[width=0.1\linewidth]{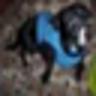}}
			\subfloat[]{\includegraphics[width=0.1\linewidth]{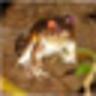}}
			\subfloat[]{\includegraphics[width=0.1\linewidth]{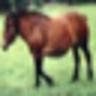}}
			\subfloat[]{\includegraphics[width=0.1\linewidth]{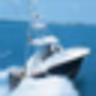}}
			\subfloat[]{\includegraphics[width=0.1\linewidth]{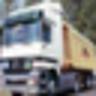}}
		\end{center}
		\caption{Sample images from CIFAR-10 database~\cite{krizhevsky2010convolutional}: (a) airplane, (b) automobile, (c) bird, (d) cat, (e) deer, (f) dog, (g) frog, (h) horse, (i) ship, and (j) truck.}
		\label{fig: Cifar-10}
	\end{figure}

\begin{figure}[htbp]
		\begin{center}
			\includegraphics[width=1\linewidth]{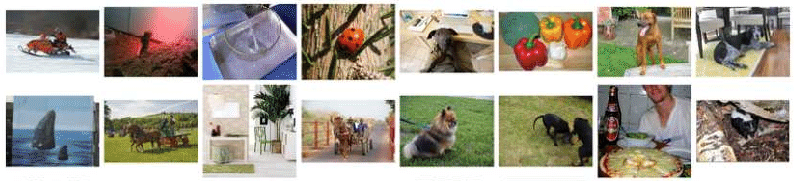}
		\end{center}
		\caption{Sample images from ImageNet-1K database~\cite{deng2009imagenet}.}
		\label{fig: ImageNet-1K}
	\end{figure}

\begin{figure}[htbp]
		\begin{center}
			\subfloat[Original ]{\includegraphics[width=0.1\linewidth]{figure/cifar10_7.jpg}}\quad
			\subfloat[Jitter 1]{\includegraphics[width=0.1\linewidth]{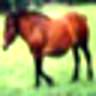}}\quad
			\subfloat[Jitter 2]{\includegraphics[width=0.1\linewidth]{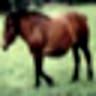}}\quad
			\subfloat[Jitter 3]{\includegraphics[width=0.1\linewidth]{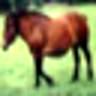}}
		\end{center}
		\caption{Color Jitter method's sample results of CIFAR-10.}
		\label{fig: Cifar-10 jitter}
	\end{figure}

In this paper, data augmentation techniques are used to create image patches from the original input. We propose a series of input images, each of which is created using different data augmentation techniques. Some examples of the input images are shown in Fig. \ref{fig: Cifar-10} and~\ref{fig: ImageNet-1K}. We use Color Jitter, which randomly changes the brightness, contrast, and saturation of images, to generate different inputs for each pod of the network. Color Jitter's sample results are shown in Fig.~\ref{fig: Cifar-10 jitter}

\subsection{Model Structure}
\begin{figure}
    \centering
    \subfloat[ResNet]{\includegraphics[width=0.25\linewidth]{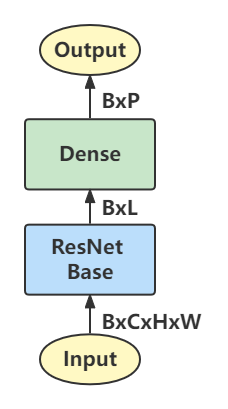}}
    \subfloat[Tripod ResNet Approach 1]{\includegraphics[width=0.38\linewidth]{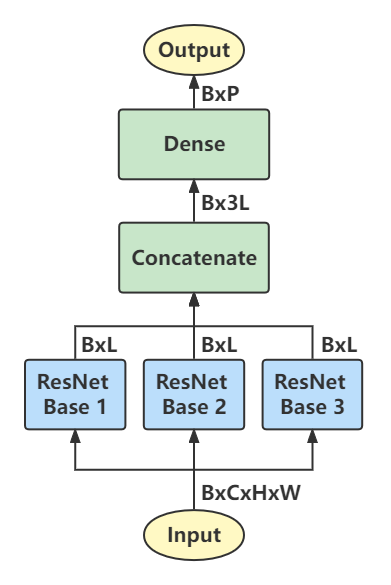}}
    \subfloat[Tripod ResNet Approach 2]{\includegraphics[width=0.38\linewidth]{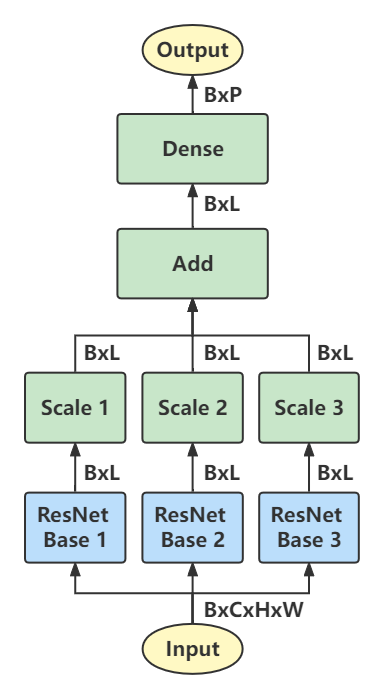}}\\
    \caption{Block diagrams of (a) a regular ResNet and two Tripod ResNet structures (b) and (c), respectively. Input images are in $B\times C\times H\times W$ (batch size, channel, height, width), and output coefficients are in $B\times P$. (a) ResNet base contains all layers up to the global average pooling layer in the regular ResNet. In (b) and (c), each ResNet base and each scale layer have their own weights. In Tripod (b), we concatenate the outputs of baseline ResNets before feeding the feature maps to the dense layer.}
    \label{fig: resnet}
\end{figure}

The proposed model consists of two or more convolutional baseline networks whose feature maps are concatenated before the fully connected layers. In this paper, we use ResNets as the baseline network since they are proven to be one of the most effective in image classification tasks. Their residual learning method allows us to go deeper into the convolution to capture better features. Different approaches such as concatenation and elementwise multiplication are also studied to combine the output feature maps in this study. We used two, three, and four ResNet networks for comparison. After deciding on the final number as three, we compare different approaches to combinations of the output feature maps. The different structures of the proposed and compared networks are given in Fig. \ref{fig: resnet}.

\section{Experimental Results}\label{sec: Experimental}
Our experiments are carried out on a Dell Workstation with an NVIDIA RTX A4000 GPU. Code is written in PyTorch in Python 3. 

\subsection{CIFAR-10 Classification Example}
Training ResNet-20 and Multiple-Pod ResNet-20s follows the implementation in ~\cite{he2016deep}. In detail, we use an SGD optimizer with a weight decay of 0.0001 and momentum of 0.9. These models are trained with a mini-batch size of 128, the initial learning rate of 0.1 for 200 epochs, and the learning rate is reduced to 1/10 at epochs 82, 122, and 163. For data argumentation, we pad 4 pixels on the training images, then do random crops to get 32 by 32 images, and then random horizontal flip images with the probability of 0.5 are applied. We normalize the images with the mean of [0.4914, 0.4822, 0.4465] and the standard variation of [0.2023, 0.1994, 02010]. During the training, the best models are saved based on the accuracy of the CIFAR-10 test dataset, and their accuracies are reported in Table~\ref{tab: CIFAR-10}. 

\begin{table}[htbp]
	\caption{Multiple Pods ResNet-20 on CIFAR-10}
    \begin{center}
    \begin{tabular}{ccc}
    \hline\noalign{\smallskip}
		Method&Parameters&Accuracy\\
        \noalign{\smallskip}\hline\noalign{\smallskip}
		ResNet-20 (official~\cite{he2016deep})&0.27M&91.25\%\\
		ResNet-20 (our implementation)&272,474&91.66\%\\
		Du-Pod ResNet-20 (approach 1)&544,938&92.25\%\\
		Tri-Pod ResNet-20 (approach 1)&817,402&92.46\%\\
		Tri-Pod ResNet-20 (approach 1 with same color jitter)&817,402&92.40\%\\
		\bf{Tri-Pod ResNet-20 (approach 1 with different color jitters)}&\bf{817,402}&\bf{92.47\%}\\
		Quad-Pod ResNet-20 (approach 1)&1,089,866&92.03\%\\
		Tri-Pod ResNet-20 (approach 2)&816,314&92.24\%\\
		ViT~\cite{ViT-CIFAR}&6.34M&90.92\%\\
		Swin Transformer~\cite{Swin_Transformer_on_CIFAR10}&4,124,352&91.20\%\\
		
    \noalign{\smallskip}\hline\noalign{\smallskip}
	\end{tabular}
\end{center}	
\label{tab: CIFAR-10}
	\end{table}

 TripodNet improved the accuracy of a single ResNet-20 as shown in Table \ref{tab: CIFAR-10}. Both the MultiPod Network with two parallel networks (Du-Pod) and QuadPod network with four parallel baseline networks improve the result of a single channel ResNet-20, however they are inferior to the TriPod Network constructed from three parallel baseline networks. 
 
 The TriPodNet shown in Fig. \ref{fig: resnet}(c) provides slightly worse results than the TriPod structure shown in Fig. \ref{fig: resnet}(b), which concatenates the $B\times L$ outputs of baseline networks before feeding the feature map to the dense layer.
 
\begin{figure}
    \centering
    \subfloat[ResNet-20]{\includegraphics[width=0.8\linewidth]{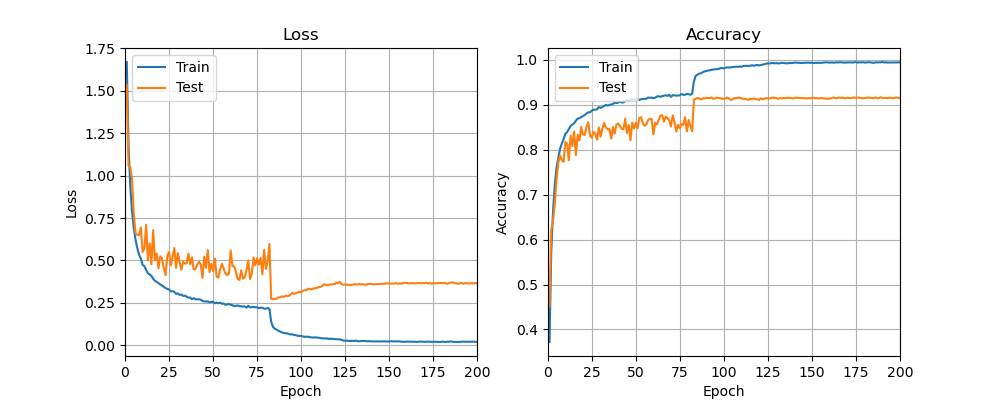}}\\
    \subfloat[Tri-Pod ResNet-20 (approach 1)]{\includegraphics[width=0.8\linewidth]{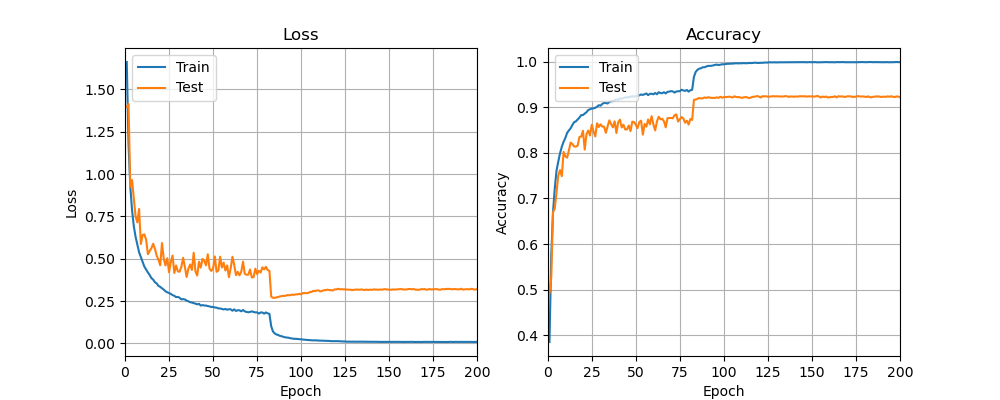}}\\
    \subfloat[Tri-Pod ResNet-20 (approach 1 with different color jitters)]{\includegraphics[width=0.8\linewidth]{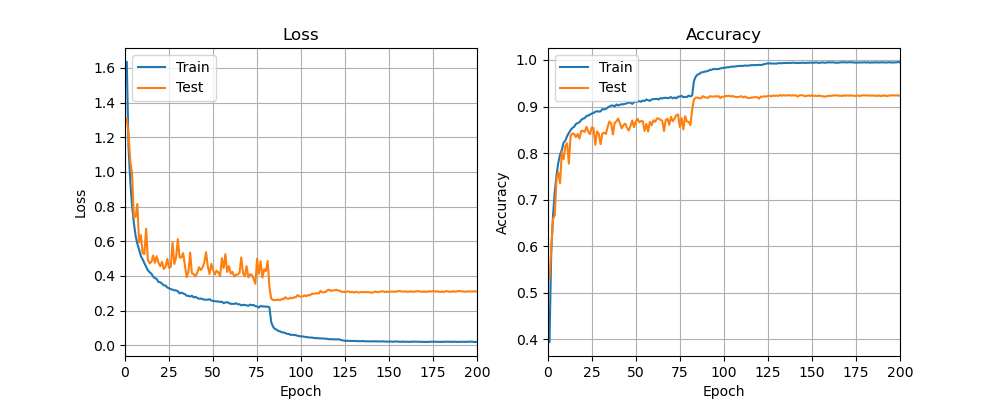}}\\
    \caption{CIFAR-10 training log.}
    \label{fig: cifar10_log}
\end{figure}

 Fig.~\ref{fig: cifar10_log} shows the training log of ResNet-20, Tri-Pod ResNet-20 (approach 1) and Tri-Pod ResNet-20 (approach 1 with different color jitters). These log figures shows that the networks are trained sufficiently because all loss plots show the over-fitting. Fig.~\ref{fig: cifar10_log} and Table~\ref{tab: CIFAR-10}show, because of the additional two pods, Tri-Pod ResNet-20 (approach 1) reaches 0.59\% higher accuracy than ResNet-20. Moreover, by applying different color jitters in the training phase, we further increase the accuracy of Tri-Pod ResNet-20 by 0.01\%. 
 
In summary, with our Tripod method, we improve the accuracy of ResNet-20 on CIFAR-10 from 91.66\% to 92.47\%. What is more, the accuracy of ViT on CIFAR-10 is reported in ~\cite{ViT-CIFAR}. The accuracy of Swin Transformer on CIFAR-10 is reported in ~\cite{Swin_Transformer_on_CIFAR10}.
TripodNet achieves better accuracy results than both ViT and Swin Transformers.

\subsection{ImageNet-1K Classification}
We employ PyTorch official ImageNet-1K training code \cite{ImageNet_training_in_PyTorch} in this section. In detail, we use an SGD optimizer with a weight decay of 0.0001 and momentum of 0.9. These models are trained with a mini-batch size of 256, the initial learning rate of 0.1 for 90 epochs, and the learning rate is reduced to 1/10 after every 30 epochs. For data argumentation, we apply random resized crops on training images to get 224 by 224 images, then random horizontal flip images with the probability of 0.5. We normalize the images with the mean of [0.485, 0.456, 0.406] and the standard variation of [0.229, 0.224, 0.225]. We evaluate the model on the ImageNet-1K validation dataset as state-of-art papers. Therefore, during the training, the best models are saved based on the center-crop top-1 accuracy on the ImageNet-1K validation dataset, and their accuracies are reported in Table~\ref{tab: ImageNet-1K}. in Table~\ref{tab: ImageNet-1K}, center-top top-5 accuracy, 10-crop top-1 accuracy, and 10-crop top-5 accuracy are provided for reference. These accuracies are from the model with the best center-crop top-1 accuracy.

\begin{table}[htbp]
	\caption{Performance of TriPod network constructed from ResNet-18 on ImageNet-1K}
    \begin{center}
    \begin{tabular}{cccccc}
    \hline\noalign{\smallskip}
        &Number of&\multicolumn{2}{c}{Center-Crop}&\multicolumn{2}{c}{10-Crop}\\
		&Parameters&Top-1&Top-5&Top-1&Top-5\\
        \noalign{\smallskip}\hline\noalign{\smallskip}
		ResNet-18 (official~\cite{he2016deep})&11,689,512&-&-&72.22\%&-\\
		ResNet-18 (our implementation)&11,689,512&69.99\%&89.45\%&72.05\%&90.81\%\\
		Tri-pod ResNet-18 (Approach 1)&35,066,536&70.89\%&89.72\%&72.93\%&90.99\%\\
		
    \noalign{\smallskip}\hline\noalign{\smallskip}
	\end{tabular}
\end{center}	
\label{tab: ImageNet-1K}
	\end{table}

TripodNet improved the accuracy of a single ResNet-18 in both Top-1 and Top-5 categories in ImageNet-1K as shown in Table \ref{tab: ImageNet-1K}.

\section{Conclusion}\label{sec: Concolusion}
In this paper, we introduce a way of increasing the number of trainable parameters of a deep learning structure by using a parallel combination of two or more convolutional networks which process the input image in parallel to achieve the same goal. By initializing the networks with different random numbers we obtained different parameters in parallel networks forming the Multipod network. We experimentally observed that the Tripod network with three parallel branches provides the best results.

In this article, pod networks are chosen as ResNets but other convolutional of fully connected feed-forward networks can be also used as pod networks.

The future work will include the construction of a MultiPod network from baseline ConvNext networks.

\bibliographystyle{IEEEtran}
\bibliography{main}
\end{document}